\PassOptionsToPackage{table}{xcolor}

\documentclass[11pt]{article}

\usepackage[final]{acl}

\usepackage{times}
\usepackage{latexsym}
\usepackage[T1]{fontenc}
\usepackage[utf8]{inputenc}
\usepackage{microtype}
\IfFileExists{inconsolata.sty}{\usepackage{inconsolata}}{}

\usepackage{amsmath, amssymb}
\usepackage{booktabs}
\usepackage{graphicx}
\usepackage{enumitem}
\usepackage{xspace}

\usepackage[most]{tcolorbox}
\newtcolorbox{finding}[1]{
  colback=blue!5!white,
  colframe=blue!50!black,
  fonttitle=\bfseries,
  title={#1},
  boxrule=0.5pt,
  arc=2pt,
  left=4pt, right=4pt, top=2pt, bottom=2pt,
  before skip=4pt, after skip=4pt,
}
\newtcolorbox{insight}[1]{
  colback=green!4!white,
  colframe=green!45!black,
  fonttitle=\bfseries,
  title={#1},
  boxrule=0.5pt,
  arc=2pt,
  left=4pt, right=4pt, top=2pt, bottom=2pt,
  before skip=4pt, after skip=4pt,
}

\newcommand{\note}[1]{}
\newcommand{\todo}[1]{}
\newcommand{\evidence}[1]{}
\newcommand{\claim}[1]{}
\newcommand{\Qwen}{Qwen2.5-VL-7B\xspace}
\newcommand{\QwenSmall}{Qwen2.5-VL-3B\xspace}
\newcommand{\QwenLarge}{Qwen2.5-VL-32B\xspace}
\newcommand{\QwenThree}{Qwen3-VL-8B\xspace}
\newcommand{\IV}{InternVL3-8B\xspace}
\newcommand{\preflmr}{PreFLMR ViT-G\xspace}
\newcommand{\infoseek}{\textsc{InfoSeek}\xspace}
\newcommand{\evqa}{\textsc{E-VQA}\xspace}
\newcommand{\codeurl}{\url{https://github.com/WeMWish/lost_at_the_end_code}}
\title{Lost at the End: Primacy Bias in Multimodal\\Retrieval-Augmented Question Answering}

\author{
  Jieyuan Liu$^{1}$ \quad Jianyang Gu$^{2}$ \quad Shijie Chen$^{2}$ \quad Jefferson Chen$^{1}$ \quad Zhen Wang$^{1}$\thanks{\,Corresponding author.} \\
  $^{1}$University of California, San Diego \quad $^{2}$The Ohio State University \\
  \texttt{\{jil029, zhw085\}@ucsd.edu} 
}

\begin{document}
\maketitle

% =============================================================================
\begin{abstract}
Knowledge-based visual question answering (KB-VQA) lets vision-language systems answer questions that exceed their parametric knowledge by conditioning a reader on passages retrieved from a Wikipedia-derived knowledge base. In pure-text long-context LLMs, retrieved-context use follows the U-shaped \emph{lost-in-the-middle} effect of \citet{liu2024lost}: information at the start and end of context is used, the middle is lost. Whether this transfers to deployed multimodal KB-VQA is open. To close this gap, we design the first controlled probe of reader-side position dependence in multimodal KB-VQA: a gold-position protocol in which only the gold passage's prompt slot varies within question. We run it on three open-source 7B/8B VLM readers and two KB-VQA benchmarks with up to 20 retrieved passages. The shape flips from U to primacy: gold-at-first beats gold-at-last by 16 to 26 points on all six combinations of reader and benchmark, an effect we call \emph{Lost at the End}; the gap holds at every scale we test, 3B to 32B, attenuating at 32B. Three targeted ablations narrow the cause. A text-only control that removes the image and changes nothing else shows the primacy is already present in text mode and does not depend on the image. Image-position and distractor-shuffle ablations trace the effect to prompt slot 0 of the instruction-tuned reader, where a second answer-bearing passage placed later is largely wasted. On a frozen reader, three retrieval-side fixes (MMR, oracle reranking, rank-based reordering) all fail to improve on the deployment default. Our findings indicate that recall@$k$ is the wrong metric for deployed KB-VQA and that the remaining headroom sits on the reader side; we release our protocol as a controlled instrument for evaluating reader-side interventions.
\end{abstract}

% =============================================================================
\section{Introduction}
\label{sec:intro}

Knowledge-based visual question answering (KB-VQA) asks fine-grained questions about entities in an image that no vision-language model reliably memorizes~\citep{lin2024preflmr,yan2024echosight,caffagni2024wikillava,cocchi2025reflectiva}. Systems therefore retrieve supporting passages and pass them to a frozen reader. When the reader is fed multiple competing passages, whether the system returns the answer depends not only on whether retrieval surfaces it but on where that passage sits in the reader's context: a passage that helps at the start of the prompt can be invisible at the end. In a deployed pipeline routing dozens of candidates per query through a frozen reader, this position dependence drives system reliability, yet the choices that optimize retrieval --- recall@$k$ targets, larger candidate pools --- implicitly assume the reader can use whatever retrieval surfaces. That assumption has not been characterized for multimodal KB-VQA. Text-only long-context QA has a well-tested answer: a U-shaped curve in which the middle of the context is underused~\citep{liu2024lost}.

Existing multimodal-RAG position-bias work has not characterized this setting. \citet{yao2025spotlight} report a U-shaped Position Sensitivity Index of 2 to 11 pp. Their benchmarks span text-only, image-only and mixed-modality comprehension; ours is KB-VQA, where the distractors are the wrong passages a retriever ranked most relevant. \citet{hu2025mrag} observe qualitatively that LVLMs prioritize early candidates on the benchmarks we use, and route around it with a top-1-only pipeline. Without controlled measurement at deployed scale, three basic questions are unanswered: is the U-shape the right model for multimodal RAG, or is the shape different? What drives the position effect, given that image proximity, retrieval similarity and prompt position are confounded in a standard pipeline? And do the obvious retrieval-side responses recover the loss?

To answer these questions, we design a gold-position protocol\footnote{Code and protocol: \codeurl} (\S\ref{sec:probe}) that holds everything bit-identical within question except the gold passage's prompt slot. Any accuracy difference must therefore come from position, and the prompts pair question-by-question for an exact bootstrap. We run it at the scale a deployed pipeline produces (top-50 \preflmr~\citep{lin2024preflmr} distractors, $k$ up to 20, frozen instruction-tuned readers with no test-time fine-tuning) on three open-source 7B/8B VLM readers (\Qwen~\citep{bai2024qwen25vl}, \IV~\citep{chen2024internvl}, \QwenThree~\citep{qwen3vl2025}) and two KB-VQA benchmarks (\infoseek~\citep{chen2023infoseek}, \evqa~\citep{mensink2023evqa}).

Running this protocol yields four findings. \emph{The shape is primacy, not U} (\S\ref{sec:main}). We use \emph{primacy bias} for the reader's systematic preference for evidence early in its prompt over otherwise-identical evidence at later positions: gold-at-first beats gold-at-last by 16 to 26 pp on all six reader-by-benchmark \emph{cells} --- three readers times two benchmarks --- with no recency rebound on five of six, and the gap holds across an order of magnitude of reader scale, from 3B to 32B parameters. Three ablations then locate the cause. \emph{The effect is a text-side property of the reader} (\S\ref{sec:text_ablation}): removing the image and changing nothing else does not shrink the gap, so primacy is not introduced by multimodal conditioning. \emph{Primacy is anchored at prompt slot 0 of the instruction-tuned reader} (\S\ref{sec:mechanism}): image-position and distractor-shuffle ablations rule out image-token proximity and retrieval similarity as primary drivers, and a multi-gold experiment shows that a second answer-bearing passage placed later contributes little once slot 0 already holds one. \emph{The obvious retrieval-side fixes do not help} (\S\ref{sec:mitigations}): MMR diversification, oracle reranking, and rank-based distractor reordering all fail to improve on the deployment default. Together these answers indicate that recall@$k$ is the wrong metric for deployed KB-VQA and that the remaining headroom sits on the reader side; we discuss what this implies for the design space of multimodal RAG in \S\ref{sec:mitigation_discussion}.

% =============================================================================
\section{Related Work}
\label{sec:related}

\noindent\textbf{Position bias in text LLMs.} The \emph{lost-in-the-middle} U-shape of \citet{liu2024lost} established that position dependence is a property of long-context reading, with follow-ups attributing it to attention-bias asymmetry~\citep{hsieh2024found}, showing direction varies by model family~\citep{hutter2025lost}, and reporting that Qwen-2.5 7B exhibits stronger positional bias than the Llama-3 family in pure-text RAG~\citep{cuconasu2025rag}. Our pattern is primacy-dominant, contrasting with \citet{liu2024lost}'s U-shape at 13B+ and reported recency at 7B; we are partially consistent with their recency component on one of six cells (\QwenThree\ $\times$ \evqa; \S\ref{sec:main}), with no recency rebound on the other five. None of these works extend to multimodal readers, where image conditioning adds an axis of confound and KB-scale retrieval changes the distractor distribution; whether the U-shape transfers, persists in a different shape, or fails entirely in multimodal KB-VQA is the question we answer.

\noindent\textbf{Multimodal position bias.} Several characterizations exist for the multimodal setting, none in deployed KB-VQA. \citet{tan2024order} and \citet{tian2025identifying} report order sensitivity in multimodal LLMs but outside KB-grounded reading; the text-side ancestor is \citet{lu2022fantastically}, who found that reordering the examples in a prompt can move accuracy from near-chance to near state-of-the-art. \citet{yao2025spotlight} report a U-shaped Position Sensitivity Index of 2 to 11 pp across MS-MARCO, ChartQA, and VEGA at 2 to 19 distractors, a text-only, an image-only, and a mixed-modality benchmark; we differ in setting (deployed KB-VQA), distractor configuration (9 deterministic top-50 distractors at controlled positions), and statistical methodology (paired-bootstrap confidence intervals, not a single point estimate). \citet{hu2025mrag} note qualitatively that LVLMs prioritize early candidates on the same KB-VQA benchmarks we use and motivate a top-1-only generation pipeline; we differ in design intent, measuring the effect and locating its mechanism instead of routing around it. \citet{wu2025posbias} concurrently characterize position bias on the retriever side, a separate question from the reader-side effect we study. Closest in spirit is \citet{wang2024mmniah}, whose MM-NIAH benchmark inserts synthetic text and image content into long real interleaved documents and tests whether the model can find it, reporting that a lost-in-the-middle pattern is present in MLLMs. We differ in what supplies the context and in what varies. They insert synthetic content into real documents and vary its depth and the document's length. Ours is the actual top-50 output of a strong KB-VQA retriever, and we vary only the gold passage's position, with distractor identity and ordering held bit-identical within question. The two are complementary, and their setting is a natural next place to run our protocol (Limitations).

% =============================================================================
\section{Probing Protocol}
\label{sec:probe}

\begin{figure*}[!t]
\centering
\includegraphics[width=\textwidth]{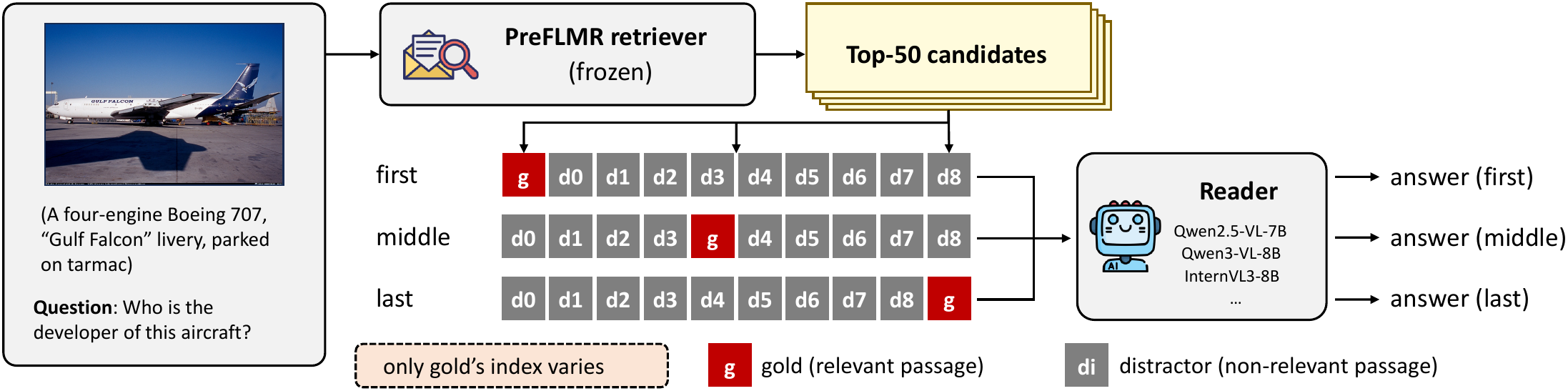}
\caption{\textbf{Pipeline and position-permutation protocol.} The image and question are encoded by a frozen \preflmr\ retriever, which returns 50 candidate passages. From these we construct three 10-passage prompts that are bit-identical except for the index of the gold passage \(g\) (red): \emph{first} (index 0), \emph{middle} (index 4), \emph{last} (index 9). Nine distractors \(d_0,\ldots,d_8\) (grey) are identical across cells per question. A frozen reader (\Qwen, \IV, or \QwenThree) generates one answer per cell.}
\label{fig:pipeline}
\end{figure*}

The probe builds on a standard frozen KB-VQA pipeline (Figure~\ref{fig:pipeline}): a multi-vector retriever produces top-50 candidate passages from a fixed knowledge base, and a vision-language reader generates an answer from the image, the question, and a $k$-passage subset of those candidates. No component is fine-tuned. Three commitments shape the design choices below: the reader's response to gold position must be isolated from retrieval rank, distractor identity, image conditioning and scoring noise; every effect must hold at deployment scale; and every measurement must be cross-replicable across reader families, benchmarks, and the scoring matchers a deployer might apply.

\noindent\textbf{Deployment-scale instantiation.} We evaluate on the two canonical knowledge-based VQA benchmarks: \infoseek~\citep{chen2023infoseek}, with a 98K-passage knowledge base, and \evqa~\citep{mensink2023evqa}, with a 51K-passage knowledge base, both drawn from the M2KR release~\citep{lin2024preflmr}. From each we draw a seed-locked subset of 500 question ids; \infoseek\ contributes 496 after excluding four questions with too few distractors, \evqa\ contributes 500. These knowledge bases are Wikipedia-derived and contain tens of thousands of passages; we scope our claims to that regime and do not extrapolate to corpora orders of magnitude larger. Retrieval uses \preflmr~\citep{lin2024preflmr}, a multi-vector ColBERT-style retriever~\citep{khattab2020colbert} that reports state-of-the-art results on M2KR; we take its top-50 candidates per query. The three readers are \Qwen~\citep{bai2024qwen25vl}, \IV~\citep{chen2024internvl}, and \QwenThree~\citep{qwen3vl2025}, run frozen and decoded greedily with a short-answer length limit, all sharing the same fixed system prompt and numbered passage block (App.~\ref{app:prompts}). Precision, library versions and compute are documented in App.~\ref{app:compute}.

\noindent\textbf{Within-question position protocol.} The probe itself is a within-question paired design. For each question we construct three $k$-passage prompts that are bit-identical except for the index of the gold passage: it occupies slot $0$ in the \emph{first} cell, slot $\lfloor (k-1)/2 \rfloor$ in the \emph{middle} cell, and slot $k-1$ in the \emph{last} cell. The other $k-1$ slots are filled with distractors that are identical in identity and ordering across the three cells per question (App.~\ref{app:distractor}). Each question therefore yields an exactly matched set of first, middle, and last results, which is what the paired bootstrap resamples. The gold passage is the highest-ranked positive in the \preflmr\ top-50; when no top-50 candidate is positive, we fall back to the dataset's first-listed positive for that question (the case for 142 of the 496 \infoseek\ questions; we report a fallback-gold sensitivity analysis in App.~\ref{app:fallback_sens}). The distractor pool is the remaining top-50 non-positives: the hardest, topically near-gold negatives a strong retriever would surface in deployment.

\noindent\textbf{Scoring and statistics.} For \infoseek\ we port the evaluation code released alongside \citet{chen2023infoseek} by its first author, which the benchmark repository links but labels community-contributed. \evqa\ does have a specified metric: \citet{mensink2023evqa} define evaluation in terms of BEM~\citep{bulian2022bem}, a learned answer-equivalence model, and the M2KR downstream setup of \citet{lin2024preflmr} adopts the same choice. What the M2KR release we consume does \emph{not} ship is a scorer implementation or a BEM checkpoint. We therefore report three lexical matchers (normalized substring, word-boundary, exact equality), treat substring as canonical, and give sensitivity in App.~\ref{app:scoring}; both \evqa\ verdicts we re-score hold under all three. Lexical matching is stricter than a learned equivalence model, so our gaps are conservative: relaxing \infoseek\ scoring to lenient containment \emph{widens} the gap (App.~\ref{app:scoring}). Our text-only NQ-Open control runs (App.~\ref{app:p27_text_internvl3}, App.~\ref{app:p29_text_qwen3vl}) use \citeauthor{liu2024lost}'s released scoring code on a seed-locked sample of 500 of their 2{,}655 paragraph-answer queries. Per-cell accuracy uses Wilson 95\% intervals; between-cell deltas use a paired bootstrap ($B{=}10{,}000$, fixed seed), and a delta is \emph{separable} when its interval excludes zero. Effect-size verdicts come from a five-way classifier whose numeric bands were fixed in a pre-registration note before each experiment it covers (App.~\ref{app:stats}). Our subsets match the full released splits on all gated covariates (App.~\ref{app:subsets}).

% =============================================================================
\section{Experiments}
\label{sec:insights}

\subsection{Primacy, Not U-Shape}
\label{sec:main}

\begin{finding}{Finding 1: Primacy, not U-shape}
Gold-at-first beats gold-at-last by 16 to 26 pp on all six combinations of reader and benchmark; middle and last sit close together, far below first. The shape is primacy-dominant, not the U reported for text-only long-context LLMs.
\end{finding}

\begin{figure*}[t]
\centering
\includegraphics[width=\textwidth]{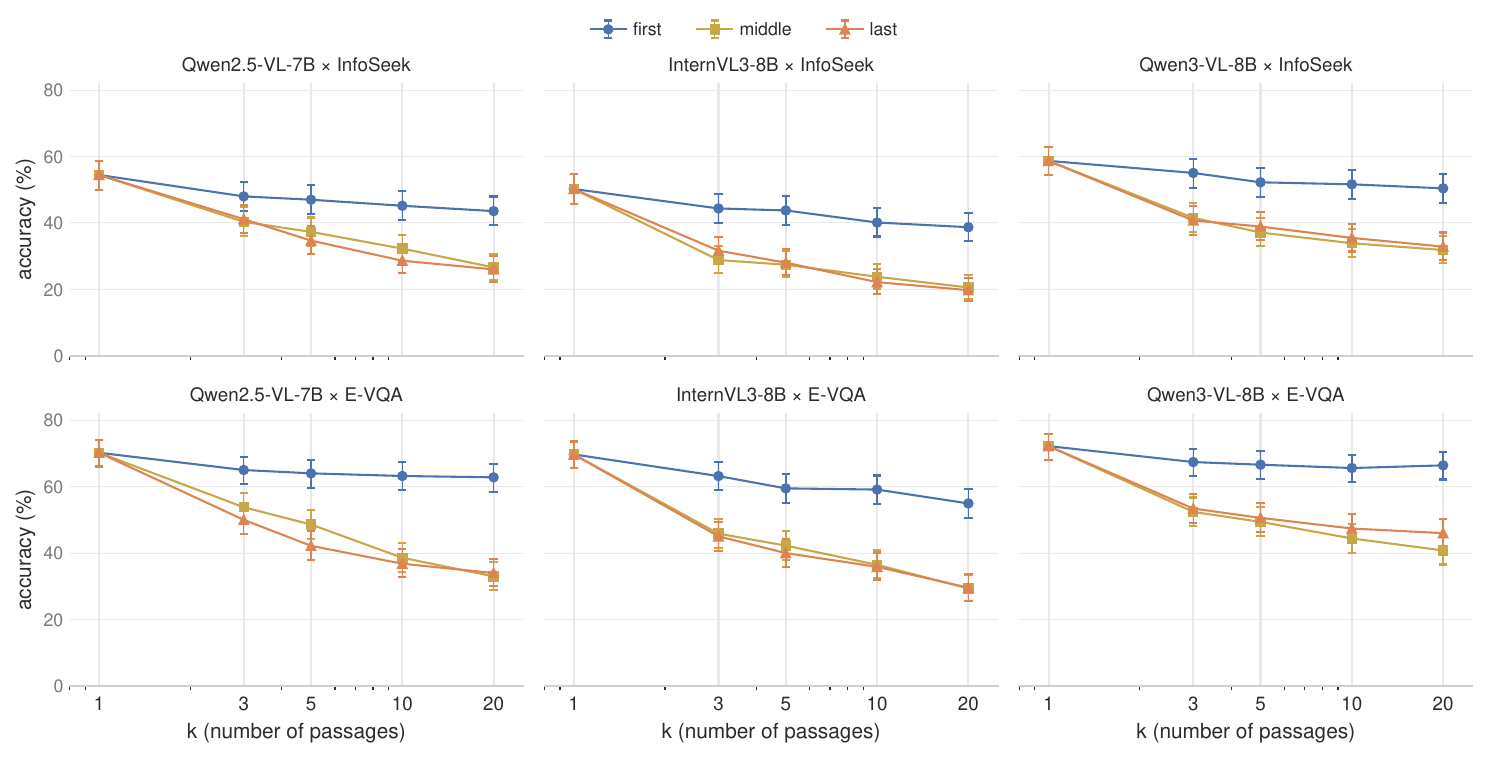}
\caption{\textbf{Position effect across $k$ on six reader$\times$benchmark cells.} Top row: \infoseek\ (\Qwen, \IV, \QwenThree, left to right); bottom row: \evqa\ (same column ordering). \emph{First} decays gracefully on all six panels; \emph{middle} and \emph{last} drop steeply and collapse together below it. The middle-vs-last ordering on \QwenThree\ \evqa\ separably inverts at higher $k$ (last $>$ middle by $5.2$ pp at $k{=}20$, paired-bootstrap CI $[+2.6, +8.0]$); the other five panels show no recency rebound at $k{=}20$. Error bars are Wilson 95\% CIs.}
\label{fig:p29_6panel}
\end{figure*}

We first establish the effect on \Qwen\ reading top-10 \infoseek\ passages, varying only the gold passage's slot among \emph{first}, \emph{middle}, and \emph{last} (Figure~\ref{fig:pipeline}). Accuracy drops sharply with gold position: 45.2\% at first, 32.3\% at middle, 28.6\% at last ($n{=}496$; Wilson CIs as error bars in Figure~\ref{fig:p29_6panel}). The first-vs-last gap of $+16.5$ pp serves as the comparator anchor below. Restricting to the 354 questions with a gold in the \preflmr\ top-50 raises absolute accuracy by $2.9$ to $4.1$ pp across the three $k{=}10$ cells (and by at most $5.2$ pp across the full $k$-sweep) but preserves the shape (App.~\ref{app:fallback_sens}).

\noindent\textbf{Shape across $k$ on all six cells.} Figure~\ref{fig:p29_6panel} sweeps $k \in \{1, 3, 5, 10, 20\}$ on all six reader-by-benchmark combinations. At $k{=}1$ all three conditions are the same prompt (50--72\%; Table~\ref{tab:p26_full}): even with gold alone the reader is the binding constraint. As $k$ grows every panel separates the same way --- \emph{first} decays gracefully while \emph{middle} and \emph{last} drop steeply and collapse together below it. We state the pattern as ``first $\gg$ middle $\approx$ last'' rather than the stricter ``first $>$ middle $>$ last'' because the middle-vs-last ordering is reader- and dataset-specific and can flip within CI noise (App.~\ref{app:p26_ksweep}, App.~\ref{app:p29_qwen3vl}). The first-vs-last gap stays large and same-signed on all six. \emph{No recency rebound} appears at $k{=}20$ on five of six cells; the sixth is the exception discussed next.

\noindent\textbf{The one exception: a sign reversal on \QwenThree\ $\times$ \evqa.} On this single panel, last sits above middle --- by $3.0$ pp at $k{=}10$ and $5.2$ pp at $k{=}20$, both separable (App.~\ref{app:p29_qwen3vl}). The reader is not simply recency-biased: in text-only NQ-Open it shows the \emph{opposite} tail (middle $>$ last by $3.6$ pp, separably; App.~\ref{app:p29_text_qwen3vl}). We do not commit to a mechanism; candidates include \QwenThree's architecture (DeepStack encoder, interleaved-MRoPE, longer pretraining context) and \evqa's longer answer surface. The first-vs-last gap here remains large ($+18.2$ pp at $k{=}10$ and $+20.4$ pp at $k{=}20$), so the reversal sits on top of the primacy story rather than in place of it, and it is partially consistent with the recency component \citet{liu2024lost} report at text-LLM scale.

\noindent\textbf{Cross-cell generalization.} Table~\ref{tab:replication} reports the first/middle/last protocol at $k{=}10$ on all six reader-by-benchmark combinations. Every first-vs-last delta is large, same-signed, and separable from zero, ranging from $+16.1$ pp (\QwenThree\ \infoseek) to $+26.4$ pp (\Qwen\ \evqa); all six fall within a factor of two of the anchor (App.~\ref{app:stats}). The two largest are \Qwen\ and \IV\ on \evqa\ ($1.6$ and $1.4\times$). Exact-match scoring brings both within $3$ pp of the anchor (App.~\ref{app:scoring}), so \evqa's free-form answers widen them through our substring matcher.

\noindent\textbf{Scale: the gap holds from 3B to 32B, but attenuates at 32B.} Reader capacity is the obvious alternative explanation --- a stronger reader might simply use late evidence better. We hold the protocol fixed and swap only the reader, adding \QwenSmall\ and \QwenLarge\ on \infoseek\ at $k{=}10$ (Table~\ref{tab:scale}). The gap is separable at every scale but does not shrink steadily with it: $+14.5$ pp at 3B, $+16.5$ pp at 7B, $+10.5$ pp at 32B. The 32B gap sits separably below the 7B anchor (paired gap-of-gaps $-6.05$ pp $[-10.28, -2.01]$), whereas 3B-vs-7B is not separable ($-2.02$ pp $[-6.25, +2.22]$), so we claim attenuation only at 32B. The 32B reader is more accurate at every position, but it gains $+8.3$ and $+8.7$ pp at middle and last against only $+2.6$ pp at first: scale helps the later slots more than the first, narrowing the gap without closing it. The failure signature of \S\ref{sec:mechanism} is unchanged: distractor-lift errors (answers taken from a distractor rather than the gold) still grow with gold position ($20.8 \to 37.6 \to 41.8\%$) and the chosen distractor still sits at median \preflmr\ rank 1--2. Scale is therefore a partial and expensive mitigation, not a fix. The 32B row is \infoseek-only: the matching \evqa\ run reached $77.4$ GB of peak allocation against the $75$ GB ceiling we set on an 80 GB H100 and was aborted after 50 of 1{,}500 prompts. We declined to quantize or reduce image resolution, since either would introduce a precision confound absent from every other cell, so we report the 32B result on one benchmark.

\begin{table}[t]
\centering
\footnotesize
\setlength{\tabcolsep}{4pt}
\caption{\textbf{Reader scale on \infoseek\ at $k{=}10$ ($n{=}496$).} Protocol identical across rows; only the reader changes. Accuracies in \%, $\Delta$ in unrounded pp with paired-bootstrap 95\% CIs. Gap-of-gaps against the 7B anchor is $-2.02$ $(-6.25,+2.22)$ at 3B (not separable) and $-6.05$ $(-10.28,-2.01)$ at 32B (separable).}
\label{tab:scale}
\resizebox{\columnwidth}{!}{%
\begin{tabular}{lrrr>{\columncolor{yellow!18}}l}
\toprule
Reader & first & middle & last & $\Delta$(first$-$last) \\
\midrule
\QwenSmall                    & 40.7 & 26.6 & 26.2 & $+14.5$ {\scriptsize $(+10.9,+18.1)$} \\
\Qwen\ {\scriptsize (anchor)} & 45.2 & 32.3 & 28.6 & $+16.5$ {\scriptsize $(+12.7,+20.6)$} \\
\QwenLarge                    & 47.8 & 40.5 & 37.3 & $+10.5$ {\scriptsize $(+7.5,+13.9)$}  \\
\bottomrule
\end{tabular}}
\end{table}

\noindent\textbf{Robustness.} The re-scored \evqa\ verdicts hold under the word-boundary and strict-equality matchers; excluding fallback-gold rows shifts deltas by at most 1.83 pp; multi-answer scoring flips zero rows. Vanilla \preflmr\ top-10 (no gold injection) gives 43.0\% \evqa\ / 29.8\% \infoseek\ for \Qwen, in range of published M2KR systems.

\begin{insight}{Insight 1: Distribute training answers across positions}
VLM instruction-tuning corpora should include multi-passage retrieval examples with gold answers spread across all positions, not concentrated in the single-passage or top-1 format that dominates current recipes~\citep{liu2023llava,dai2023instructblip}.
\end{insight}

\begin{table}[t]
\centering
\footnotesize
\setlength{\tabcolsep}{3.5pt}
\caption{\textbf{Cross-benchmark generalization of the first-vs-last position effect at $k{=}10$.} Paired-bootstrap 95\% CIs use $10{,}000$ resamples and a fixed seed; all six deltas fall in the same-magnitude band against the anchor. Per-cell accuracies are rounded to 1 dp; $\Delta$ is computed from unrounded values and may differ from displayed-value subtraction by up to 0.1 pp.}
\label{tab:replication}
\resizebox{\columnwidth}{!}{%
\begin{tabular}{llcc>{\columncolor{yellow!18}}cll}
\toprule
Reader & Benchmark & first (\%) & last (\%) & $\Delta$ (pp) & {\scriptsize 95\% CI} & ratio \\
\midrule
\Qwen      & \infoseek\ ($n{=}496$) & 45.2 & 28.6 & $+16.5$ & {\scriptsize $(+12.7,+20.6)$}                                    & $1.00\times$ (anchor) \\
\Qwen      & \evqa\ ($n{=}500$)     & 63.2 & 36.8 & $+26.4$ & {\scriptsize $(+22.2,+30.6)$}      & $1.60\times$ \\
\IV        & \infoseek\ ($n{=}496$) & 40.1 & 22.2 & $+17.9$ & {\scriptsize $(+14.1,+22.0)$}      & $1.09\times$ \\
\IV        & \evqa\ ($n{=}499$)     & 59.1 & 35.9 & $+23.2$ & {\scriptsize $(+19.0,+27.5)$}      & $1.41\times$ \\
\QwenThree & \infoseek\ ($n{=}496$) & 51.6 & 35.5 & $+16.1$ & {\scriptsize $(+12.5,+20.0)$}      & $0.98\times$ \\
\QwenThree & \evqa\ ($n{=}500$)     & 65.6 & 47.4 & $+18.2$ & {\scriptsize $(+14.2,+22.2)$}      & $1.10\times$ \\
\bottomrule
\end{tabular}}
\end{table}

\subsection{Primacy Is a Text-Side Property}
\label{sec:text_ablation}

\begin{finding}{Finding 2: Removing the image does not shrink the gap}
On matched KB-VQA data with only the image removed, the first-vs-last gap is at least as large as in the multimodal setting on all three reader-by-benchmark cells tested. Primacy is a property of the reader's text-position prior; multimodal conditioning does not introduce it.
\end{finding}

How much of the multimodal effect is inherited from the reader's underlying text-mode behavior? A text-only ablation on a \emph{different} corpus cannot answer this: it varies corpus and modality together, leaving the difference unattributable to either. We therefore hold the corpus fixed and vary only whether the reader receives the image. A cross-corpus amplification claim made in an earlier version is withdrawn for this reason (Limitations).

\noindent\textbf{Matched-corpus control.} We re-run three reader-by-benchmark cells holding everything fixed except whether the reader receives the image: same M2KR questions, same gold passage, same distractors in the same slots, same decoding, same scorer. Prompts are bit-identical to the multimodal runs minus the image block, and no retrieval is repeated. Removing the image does not shrink the first-vs-last gap on any cell; the point estimate grows on all three (Table~\ref{tab:text_ablation}).

\noindent\textbf{The direction is consistent; the interaction separates on one cell of three.} We test the difference of gaps directly, resampling questions jointly across all four conditions. It is separable only for \Qwen\ $\times$ \infoseek\ ($+6.05$ pp $[+1.81, +10.28]$); \Qwen\ $\times$ \evqa\ ($+3.40$ $[-0.80, +7.60]$) and \IV\ $\times$ \infoseek\ ($+1.21$ $[-3.02, +5.44]$) straddle zero. This supports the claim that primacy does not depend on the image. The stronger claim that the image \emph{dampens} primacy would require the interaction to separate on more than one cell.

\noindent\textbf{Where the image acts.} On \Qwen\ $\times$ \infoseek\ the cost of removing the image falls very unevenly across positions --- \emph{middle} loses $15.12$ pp and \emph{last} $10.08$ pp, against only $4.03$ pp at \emph{first} --- and the image-at-end ablation of \S\ref{sec:mechanism} breaks a near-subset of the very same questions, suggesting the two manipulations are graded doses of one mechanism in which visual grounding partly protects evidence outside slot 0. This is an \infoseek-specific hypothesis, not an established cause: on \evqa\ every per-position difference straddles zero. Full per-position intervals and the row-level overlap statistics are in App.~\ref{app:vision_protection}.

\noindent\textbf{Primacy is present in a different corpus too.} All three readers also show first-vs-last primacy on the NQ-Open text-only protocol ($+7.4$, $+5.2$, $+6.4$ pp; App.~\ref{app:p27_text_internvl3}, App.~\ref{app:p29_text_qwen3vl}), evidence that the effect is not an artifact of one corpus.

\begin{table}[t]
\centering
\footnotesize
\setlength{\tabcolsep}{4pt}
\caption{\textbf{Matched-corpus text-only control.} Same questions, same gold passage, same distractors in the same slots, same decoding and same scorer; the only change is whether the reader receives the image. Per-cell accuracies in \%; $\Delta$ in unrounded pp. Removing the image does not shrink the gap on any cell. The paired-bootstrap difference between the two gaps, computed over the shared question set, is $+6.05$ $(+1.81, +10.28)$ for \Qwen\ $\times$ \infoseek, $+3.40$ $(-0.80, +7.60)$ for \Qwen\ $\times$ \evqa, and $+1.21$ $(-3.02, +5.44)$ for \IV\ $\times$ \infoseek: same-signed on all three cells, separable on one.}
\label{tab:text_ablation}
\resizebox{\columnwidth}{!}{%
\begin{tabular}{llrrr>{\columncolor{yellow!18}}l}
\toprule
Reader $\times$ Benchmark & Image & first & middle & last & $\Delta$(first$-$last) \\
\midrule
\Qwen\ $\times$ \infoseek\ {\scriptsize($n{=}496$)} & yes & 45.2 & 32.3 & 28.6 & $+16.5$ \\
                                                    & no  & 41.1 & 17.1 & 18.5 & $+22.6$ {\scriptsize $(+18.5,+26.6)$} \\
\midrule
\Qwen\ $\times$ \evqa\ {\scriptsize($n{=}500$)}     & yes & 63.2 & 38.6 & 36.8 & $+26.4$ \\
                                                    & no  & 66.0 & 36.6 & 36.2 & $+29.8$ {\scriptsize $(+25.4,+34.2)$} \\
\midrule
\IV\ $\times$ \infoseek\ {\scriptsize($n{=}496$)}   & yes & 40.1 & 23.8 & 22.2 & $+17.9$ \\
                                                    & no  & 37.1 & 17.1 & 17.9 & $+19.2$ {\scriptsize $(+15.5,+23.0)$} \\
\bottomrule
\end{tabular}}
\end{table}

\noindent\textbf{Scope and reader-family direction.} Primacy is not a multimodal-RAG artifact in these readers. Position-bias \emph{direction} at the 7B--8B scale is nonetheless reader-family-specific: \citet{liu2024lost} report Llama-2-7B as solely recency-biased on the same NQ protocol, while all three readers we test are primacy-biased, and \citet{cuconasu2025rag} independently report that Qwen-2.5 7B exhibits strong positional bias in text RAG. We therefore scope the text-side claim to the open-source instruction-tuned VLM families we tested; whether it holds for Llama-3-based VLMs or proprietary models is open (Limitations).

\begin{insight}{Insight 2: Mitigate on the text side, not the vision side}
The gap survives with no visual input and anchors on a text slot, so mitigation should target the reader's text-position prior, not the multimodal interface. Position-aware fine-tuning, data rebalancing and attention calibration developed against text-LLM primacy~\citep{hsieh2024found,xiao2024streaming} are the direct candidates to transfer.
\end{insight}

\subsection{Primacy Is Anchored at Prompt Slot 0}
\label{sec:mechanism}

\begin{finding}{Finding 3: Anchored at prompt slot 0}
Primacy is anchored at prompt slot 0 of the instruction-tuned reader. Two ablations rule out image-token proximity and retrieval similarity as primary drivers, and a multi-gold test shows evidence placed after slot 0 is largely wasted. The underlying mechanism~\citep{su2024roformer,press2022alibi,xiao2024streaming} remains open (Limitations).
\end{finding}

What is the reader doing wrong at later gold positions, and which axis drives it? A typical \infoseek\ case: asked who operates Dresden Hauptbahnhof with gold at slot 9, the reader returns a Munich U-Bahn operator, lifted from the \preflmr\ rank-1 distractor about Munich (further samples ship with the code; App.~\ref{app:qualitative}). The pattern is general --- when the reader is wrong, its answer is most often substring-present in whichever passage sits at slot 0 or 1.

\noindent\textbf{Failure-bucket diagnostic.} We bucket every error in the anchor cells by string-relationship to the prompted passages: \emph{extraction-failed} (prediction in gold only), \emph{distractor-lift} (in a distractor but not gold), \emph{ambiguous} (both), \emph{hallucinated} (neither). Distractor-lift grows monotonically with gold position ($30.1 \to 53.3 \to 56.5\%$) while extraction-failed drops ($36.4 \to 17.9 \to 13.8\%$), the other two buckets staying roughly constant (App.~\ref{app:bucket_renormalize}): the modal failure flips from ``did not use the right passage'' to ``used a wrong passage instead.'' Among distractor-lift rows the chosen distractor's median \preflmr\ rank is $\leq 1$ in every cell, which under rank-order placement means the first or second distractor slot. Two readings fit: a \emph{position} story (the reader locks onto slot 0) and a \emph{similarity} story (it locks onto rank-1 wherever placed). The next two ablations rule out proximity and similarity in turn.

\noindent\textbf{Image-at-end ablation rejects proximity as primary.} If image-token proximity to early passages drove primacy, moving the image after the passage block should shrink or invert the gap. Pre-registered rule: a delta at or above $0.75\times$ the anchor rejects proximity. Result on the anchor cell ($n{=}496$): first-vs-last $+18.55$ pp, $1.12\times$ the anchor. Proximity is therefore not the primary driver. A post-hoc middle-vs-last sign flip does emerge --- $\Delta(\text{middle}-\text{last})$ inverts from $+3.6$ to $-3.6$ pp, separably (App.~\ref{app:image_end_posthoc}) --- so proximity plays a secondary modulating role on adjacent slots.

\noindent\textbf{Distractor-shuffle ablation rejects similarity.} The position- and similarity-anchoring readings are confounded in the default protocol because the rank-1 distractor sits at slot 0. We dissociate them with a controlled shuffle on the gold-at-last cell ($n{=}496$, \infoseek, \Qwen): \emph{Cell A} (rank-order baseline; rank-1 at slot 0), \emph{Cell B} (reverse-rank; rank-1 at slot 8, rank-9 at slot 0), \emph{Cell C} (per-question shuffle; rank-1 at a random slot). Table~\ref{tab:shuffle} and Figure~\ref{fig:distractor_shuffle} cross-tabulate distractor-lift errors by (i) which slot's passage was chosen and (ii) that passage's PreFLMR rank. In the dissociable cells B and C, slot-0 selection runs at $3.83\times$ chance ($1/9$) while rank-1 selection runs at $1.01\times$ (App.~\ref{app:p23_shuffle_fig}). Cell B is decisive: with the least-similar (rank-9) distractor at slot 0, the reader picks it 73 times; with the most-similar (rank-1) distractor at slot 8, it picks just 7. The reader follows the slot, not the similarity.

\begin{table}[t]
\centering
\footnotesize
\caption{\textbf{Distractor-shuffle cross-tabulation.} In dissociable cells B+C, slot-0 selection runs well above the $1/9$ chance baseline while rank-1 selection sits at chance. Mechanism replicates on \evqa.}
\label{tab:shuffle}
\resizebox{\columnwidth}{!}{%
\begin{tabular}{llrrr}
\toprule
Cell & rank-1 at & $n$ (dl) & slot 0 & rank 1 \\
\midrule
\multicolumn{5}{l}{\textit{\Qwen\ $\times$ \infoseek\ ($n{=}496$)}} \\
A (rank-order) & slot 0 & 199 & 0.558 & 0.558 \\
B (reverse)    & slot 8 & 190 & 0.384 & 0.037 \\
C (shuffle)    & random & 208 & 0.466 & 0.188 \\
\rowcolor{yellow!18} B+C mean       &        &     & 0.425\,{\scriptsize ($3.83\times$)} & 0.112\,{\scriptsize ($1.01\times$)} \\
\midrule
\multicolumn{5}{l}{\textit{\Qwen\ $\times$ \evqa\ ($n{=}500$)}} \\
A (rank-order) & slot 0 & 213 & 0.601 & 0.601 \\
B (reverse)    & slot 8 & 217 & 0.521 & 0.055 \\
C (shuffle)    & random & 197 & 0.462 & 0.168 \\
\rowcolor{yellow!18} B+C mean       &        &     & 0.491\,{\scriptsize ($4.42\times$)} & 0.111\,{\scriptsize ($1.00\times$)} \\
\bottomrule
\end{tabular}}
\end{table}

The mechanism replicates on \evqa\ (Table~\ref{tab:shuffle}, lower panel): slot-0 selection runs at $4.42\times$ baseline while rank-1 sits at exactly $1.00\times$. On neither benchmark does any shuffle variant separably improve on the deployment default (App.~\ref{app:shuffle_operational_audit}), so this is a mechanism result, not a recommended intervention.

\noindent\textbf{Slot-0 privilege, not decay toward the end.} ``Lost at the end'' and ``slot 0 is privileged'' predict the same first-vs-last gap but differ on what a \emph{second} correct passage is worth. We separate them with a pre-registered multi-gold experiment on the 240 \infoseek\ questions having at least two top-50 positives that both contain the answer (rules and thresholds fixed before execution; App.~\ref{app:methodology}). Six cells place golds at slots $\{0\}$, $\{4\}$, $\{9\}$, $\{0,9\}$, $\{4,9\}$, $\{0,4\}$, with the remaining slots filled from the same ranked distractor list; since $k{=}10$ is fixed, a two-gold cell holds eight of the nine distractors used in the one-gold cells (Table~\ref{tab:multigold}). A second gold at slot 9 is worth $+2.92$ pp $[-0.42, +6.25]$ when slot 0 already holds one (C4$-$C1) --- not separable --- and $+4.17$ pp $[+1.25, +7.50]$ when the first gold sits at slot 4 instead (C5$-$C2). The pre-registered contrast C5$-$C3 gives $+7.08$ pp $[+3.33, +10.83]$, which also promotes the first gold from slot 9 to slot 4. One gold at slot 0 also beats two at $\{4,9\}$ by $+9.17$ pp $[+4.17, +14.58]$. The registered rule ordered C4$-$C1 below C5$-$C3, which holds and is robust (McNemar $p{=}0.14$ vs $p{=}0.0002$); a post-hoc test of their \emph{difference} gives $+4.17$ pp $[+0.42, +8.33]$, which we flag as fragile rather than headline, since $p \approx 0.04$, its lower bound sits one question above zero, and it fails a two-test Bonferroni adjustment. We report the null as a bound, not a zero: with a minimum detectable effect of $4.83$ pp at $n{=}240$, a later redundant gold contributes under $6.25$ pp against a $+16.25$ pp gap on the same questions. Failure buckets agree --- a gold at slot 0 cuts distractor-lift from 85 errors to 22 --- and of the 129 correct answers under golds at $\{0,9\}$, only 6 are uniquely attributable to the slot-9 passage. We therefore adopt the slot-0 framing: the reader privileges the first passage slot, and what follows is under-used whether it is a distractor or a second copy of the answer.

\noindent\textbf{What this experiment does not show.} The two golds are nearly always different chunks of one Wikipedia article (240 of 240 share a title; median content token-Jaccard $0.16$; two pairs are byte-identical, App.~\ref{app:multigold_detail}), the M2KR \infoseek\ split supplies \emph{pseudo}-gold passages rather than curated oracles, and \evqa\ as distributed in M2KR has no multi-positive questions at all. This tests positional privilege under \emph{redundant} evidence, not multi-hop reasoning where two distinct passages must be combined; App.~\ref{app:multigold_detail} gives the full construction and its caveats.

\begin{table}[t]
\centering
\footnotesize
\setlength{\tabcolsep}{5pt}
\caption{\textbf{Multi-gold cells} (\Qwen\ $\times$ \infoseek, $k{=}10$, $n{=}240$, pre-registered). Each cell places one or two answer-bearing gold passages at the listed slots; the remaining slots are filled from the same ranked distractor list, so a two-gold cell holds eight of the nine distractors used in the one-gold cells. A second gold separably helps only when slot 0 is not already gold. Construction and caveats: App.~\ref{app:multigold_detail}.}
\label{tab:multigold}
\begin{tabular}{llr@{\hskip 14pt}llr}
\toprule
Cell & golds @ & acc (\%) & Cell & golds @ & acc (\%) \\
\midrule
C1 & $\{0\}$ & 50.8 & C4 & $\{0,9\}$ & 53.8 \\
C2 & $\{4\}$ & 37.5 & C5 & $\{4,9\}$ & 41.7 \\
C3 & $\{9\}$ & 34.6 & C6 & $\{0,4\}$ & 51.3 \\
\bottomrule
\end{tabular}
\end{table}

\begin{insight}{Insight 3: Rerank for rank 1, not recall@$k$}
Reranker training should explicitly reward placing the correct passage at rank 1 and penalize placing it anywhere else; recall@$k$ objectives reward the wrong target here, since a frozen reader privileges slot 0.
\end{insight}

\subsection{Retrieval-Side Fixes Don't Help}
\label{sec:mitigations}

\begin{finding}{Finding 4: Retrieval-side fixes don't help on a frozen reader}
MMR diversification, oracle reranking, and rank-based distractor reordering all fail to improve on the deployment default within paired-bootstrap CIs. With the slot-0 anchoring of \S\ref{sec:mechanism}, recall@$k$ is the wrong metric for deployed KB-VQA.
\end{finding}

All three act only on what the reader sees: diversification of the retrieved set, reranking that pushes gold toward the start of the context, and reordering of the distractors themselves. We focus on training-free retrieval-side interventions because they are the cheapest to deploy and align with the design space surveyed by \citet{hu2025mrag}. Reader-side interventions are out of scope for a frozen-reader paper but are natural next steps.

\noindent\textbf{MMR diversification does not help.} We replace top-$k$ \preflmr\ retrieval with MMR~\citep{carbonell1998mmr} over the top-50 \preflmr\ candidates, sweeping $\lambda \in \{0.3, 0.5, 0.7\}$ at $k \in \{5, 10\}$ (six cells, \infoseek, \Qwen). The pre-registered \textsc{kill} rule --- registered separately from the five-way classifier of App.~\ref{app:stats} --- abandons the intervention if the best MMR gain over vanilla is under $1$ pp at every $(k, \lambda)$. The maximum point gain is $+1.00$ pp, at the threshold rather than under it, and every paired-bootstrap CI on $\Delta(\text{MMR}-\text{vanilla})$ overlaps zero; Figure~\ref{fig:p19} in App.~\ref{app:mmr} shows the flat curves.

\noindent\textbf{Oracle reranking does not separably help.} On the gold-in-top-10 subset ($n{=}254$) we compare \emph{native} (gold at its natural \preflmr\ rank), \emph{first} (forced to index 0) and \emph{last} (index 9): accuracies 47.64\%, 50.79\%, 42.13\%. First-vs-native is $+3.15$ pp, directionally positive but not separable at this sample size; last-vs-native is $-5.51$ pp, separable on the negative side. We read this as suggestive of asymmetry rather than a characterized result --- the promotion side is underpowered at $n{=}254$ (App.~\ref{app:oracle_rerank}).

\noindent\textbf{Rank-based reordering does not help either.} Reversing the distractor ranking against the rank-order deployment default (Cells B and A of \S\ref{sec:mechanism}) gives $+2.42$ pp $[-0.60, +5.65]$ on \infoseek\ and $-3.20$ pp $[-6.60, +0.00]$ on \evqa; neither is separable (App.~\ref{app:shuffle_operational_audit}).

\begin{insight}{Insight 4: Invest in reader-side fixes, not more retrieval}
Engineering investment should shift from retrieval (larger pools, diversification, reordering) to reader-side training and inference-time intervention; no retrieval-side variant we tried improved on the deployment default.
\end{insight}

\subsection{What Reader-Side Mitigation Would Look Like}
\label{sec:mitigation_discussion}

Our results are diagnostic, so a fair question is what a practitioner should do with them. We propose no method, but the measurements constrain the design space enough to name three directions.

\noindent\textbf{Text-side calibration.} The most actionable consequence of \S\ref{sec:text_ablation} is negative: interventions aimed at the image encoder or the vision--language projection target the wrong component. The transferable candidates are those developed against text-LLM positional bias --- attention calibration that rescales per-passage attention by position-debiased relevance~\citep{hsieh2024found}, and the attention-sink account of early-token dominance~\citep{xiao2024streaming} --- both inference-time and reader-internal, matching where we locate the effect. \textbf{Rank-1 reranking objectives.} \S\ref{sec:mechanism} quantifies how little a correct passage is worth outside the front slot: with a gold already at slot 0, a second answer-bearing gold at slot 9 adds under $6.25$ pp against a $+16.25$ pp gap. An objective rewarding gold anywhere in the top $k$ optimizes a quantity the reader largely ignores; listwise permutation-aware ranking~\citep{tang2024permutation} and attention-based rerankers~\citep{chen2024icr} already operate on whole orderings rather than on passages in isolation. \textbf{Scale as a partial, costly mitigation.} Going from 7B to 32B shrinks the gap by $6.05$ pp but leaves it separable, leaves the failure signature intact, and at $k{=}10$ on \evqa\ exceeded our memory ceiling on a single 80 GB accelerator.

\noindent\textbf{Why the protocol is the contribution.} Each direction is a hypothesis our probe can adjudicate, because the gold-position design supplies what an aggregate benchmark cannot: a within-question paired contrast whose only difference is a permutation, so an intervention's effect on position dependence is measured directly rather than inferred from an accuracy delta that also moves with retrieval quality. Concurrently, \citet{jung2026bair} propose exactly the text-side calibration above. We have not evaluated it and make no priority claim, but it is the natural first candidate to run through the protocol we release.

% =============================================================================
\section{Conclusion}
\label{sec:conclusion}

This paper provides the first controlled probe of reader-side position dependence in deployed multimodal KB-VQA. Across three open-source 7B/8B readers on two benchmarks, and five models from 3B to 32B, the shape is primacy, not U; the effect survives removing the image on matched data, making it a text-side property of the reader; primacy is anchored at prompt slot 0, to the point that a second answer-bearing passage placed later is largely wasted; and three training-free retrieval-side interventions fail to improve on the deployment default. Scale attenuates the gap at 32B without closing it or changing its signature. The practical consequence is that recall@$k$ measures something the reader largely ignores, and the remaining headroom sits on the reader side. We release the protocol as a controlled instrument for evaluating any candidate reader-side or training-time intervention.

% =============================================================================
\section*{Limitations}
\label{sec:limitations}

\noindent\textbf{Scope of empirical claims.} Our findings hold within: open-source instruction-tuned VLMs from 3B to 32B (\Qwen, \IV, \QwenThree, plus \QwenSmall\ and \QwenLarge\ on \infoseek); contexts at $k \leq 20$; English KB-VQA over Wikipedia-derived bases of $10^5$ passages; greedy decoding; one image per question. We do not claim proprietary VLMs (e.g.\ GPT-4V~\citep{openai2023gpt4v}) or other prompt formats would behave the same; \citet{tian2025identifying} and \citet{yao2025spotlight} both report proprietary-vs-open-source differences. Our readers span three families, and \citet{cuconasu2025rag} report that Qwen-2.5 7B is more positionally biased than Llama-3 in text RAG, so family plausibly matters; Llama-3-based VLMs are untested here. The no-recency-rebound claim holds on five of six $k{=}20$ cells, the sixth being the \QwenThree\ $\times$ \evqa\ reversal reported in \S\ref{sec:main}, for which we commit to no mechanism.

\noindent\textbf{One task family, and a modest context budget.} Both benchmarks are single-image KB-VQA over Wikipedia-derived text, and neither requires combining two distinct passages. So we cannot separate retrieved from parametric knowledge --- for well-covered entities a reader may not use the context at all, which would attenuate rather than manufacture a position effect, but is uncontrolled here --- and $k{=}20$ is far short of what frontier readers accept, so we cannot say whether the shape persists at $10^2$--$10^3$ candidates. Long-horizon personal-memory QA with multi-gold evidence pools~\citep{mei2026atmbench} and long interleaved multimodal documents~\citep{wang2024mmniah} would address both at once, under an oracle-plus-noise construction that keeps accuracy clear of the floor where a position probe has no room to move. We ran neither, and did not test the strongest recent open readers (e.g.\ Qwen3.6-27B).

\noindent\textbf{Mechanism attribution is partial.} Section~\ref{sec:mechanism} rules out image-token proximity and retrieval similarity as primary drivers and narrows the effect to slot 0, without identifying the underlying mechanism. Instruction-tuning data is a recurring candidate: \citet[\S4.3]{liu2024lost} considered it but found the U-shape in both base and instruct MPT-30B, and \citet{lu2022fantastically} show demonstration order alone can swing in-context performance from chance to SOTA. We find the hypothesis consistent with our data but do not test it. The matched-corpus control says where \emph{not} to look --- not the vision encoder or the vision--language interface --- but a positive account needs attention- or representation-level probes we leave to future work.

\noindent\textbf{A claim withdrawn between review and publication.} The version reviewed at ARR contained a further finding: that the multimodal setting amplifies text-mode primacy by $2.2$--$4.5\times$. Reviewers correctly observed that the supporting ablation varied corpus (NQ-Open vs.\ M2KR) together with modality. We ran the prescribed matched control, it did not support the claim, and we withdrew it; \S\ref{sec:text_ablation} reports the control in its place. We record it because the reviewed version is public. What replaces it is weaker than the reverse claim would be: the control establishes that primacy does \emph{not depend on} the image, not that the image dampens it. The text-minus-multimodal interaction is same-signed on all three cells but separable on one, on \evqa\ no per-position difference separates from zero, and the control covers three reader-by-benchmark cells rather than all six (\QwenThree\ has no matched text-only run). The position-local-protection reading should be treated as an \infoseek-specific hypothesis from row-level overlap, not an established mechanism.

\noindent\textbf{Sample size and multiplicity.} Per-cell accuracies use seed-locked subsets of 486--500 questions, chosen for paired-bootstrap tractability across many cells; on smaller slices (the $n{=}254$ oracle-rerank and $n{=}240$ multi-gold cells) the noise floor is higher, and the directional first-vs-native oracle gain does not clear it. We report roughly fifty between-cell deltas without family-wise or false-discovery correction. This does not touch the headline effects --- every KB-VQA first-vs-last gap at $k{=}10$ clears zero by at least $7$ pp and would survive any standard correction, as would the null-supported claims, which correction only strengthens --- but it does confine our exposure to boundary-positive results, and we mark the one that matters (the post-hoc difference-of-increments interval in \S\ref{sec:mechanism}) as failing even a two-test Bonferroni adjustment.

\noindent\textbf{Bucket diagnostic is heuristic, and mitigations tested are training-free.} The four-bucket diagnostic is a string-match device, not a causal account: it conflates surface-form with semantic failures, cannot attribute a prediction to one source when it appears in several passages, and has paraphrase blind spots. Re-bucketing under a stricter normalizer (App.~\ref{app:bucket_renormalize}) leaves the qualitative pattern intact, but the quantitative shares should be read accordingly; it also works at passage-slot granularity, so whether the answer's location \emph{within} a passage matters is outside our scope. Separately, we do not claim that learned rerankers~\citep{yan2024echosight}, attention-based rerankers~\citep{chen2024icr}, or self-reflective selectors~\citep{cocchi2025reflectiva} cannot help --- they reshape what the reader sees in ways our oracle protocol does not test. We claim only that simple training-free retrieval-side manipulations on a frozen reader do not close the gap.

% =============================================================================
\section*{Ethical Considerations}
\label{sec:ethics}

\noindent\textbf{Potential risks.} This is a characterization study: we measure how existing frozen vision-language readers respond to ordered retrieved context, and release no new models or datasets. The findings (a primacy anchor on prompt slot 0) describe an existing property of deployed KB-VQA systems and do not expose a new attack surface; any practitioner deploying such a system can read this work and adjust their pipeline accordingly. We do not foresee significant misuse risk beyond what is inherent in publishing any analysis of deployed-system behavior.

\noindent\textbf{Artifact licenses and intended use.} All models and datasets used are publicly released, all but the 3B reader and \evqa\ under a permissive license, and our use (academic measurement of position effects on standard KB-VQA benchmarks) is consistent with each artifact's intended use. Models: \preflmr\ (MIT)~\citep{lin2024preflmr}; Qwen2.5-VL-7B-Instruct, Qwen2.5-VL-32B-Instruct and Qwen3-VL-8B-Instruct (Apache 2.0)~\citep{bai2024qwen25vl,qwen3vl2025}; InternVL3-8B (Apache 2.0)~\citep{chen2024internvl}; Qwen2.5-VL-3B-Instruct (Qwen Research License, non-commercial research use)~\citep{bai2024qwen25vl}. Datasets: M2KR benchmark (MIT) and \infoseek\ (Apache 2.0)~\citep{lin2024preflmr,chen2023infoseek}; Encyclopedic-VQA (released through the \texttt{google-research} repository, no explicit dataset LICENSE file; image content drawn from iNaturalist 2021 (CC-BY) and Google Landmarks v2 and textual content derived from Wikipedia under CC-BY-SA)~\citep{mensink2023evqa}; the NaturalQuestions-Open multi-document protocol of \citet{liu2024lost} (MIT).

\noindent\textbf{Data content.} \infoseek\ and Encyclopedic-VQA are derived from Wikipedia/Wikidata and curated visual-entity datasets. We did not perform an independent PII or offensive-content audit on top of the curation already conducted by the original dataset authors; our use is read-only and does not modify, redistribute, or augment the underlying corpora. All passages we surface to readers are taken verbatim from the M2KR-released knowledge bases.

\section*{Acknowledgements}

\noindent The authors used AI assistants (LLM-based coding and writing tools) during this project for three purposes: (i)~drafting and editing prose, (ii)~generating and debugging analysis and plotting code, and (iii)~surfacing related work for follow-up reading. All experimental design, statistical methodology, claims, and interpretations are the authors' own; all AI-suggested code was reviewed and validated against the pre-registered methodology notes (App.~\ref{app:methodology}) before producing the numbers reported here.

% =============================================================================
\bibliography{custom}

% =============================================================================
\appendix

\section{Pre-registered Methodology Notes}
\label{app:methodology}
Most experiments were preceded by a methodology note (decision rules, expected effect-size bands, comparators) committed to version control before the run. The notes that exist are: P15a (decision-rule protocol), P16 (allocator probe), P17 (gold-position anchor), P17b (second-model extension), P19 (MMR), P20 (\evqa\ replication), P20-VANILLA, P21-TEXT-LIU (text-only NQ ablation on \Qwen), P22 (image-at-end), P23 (\infoseek\ shuffle), P24 (\evqa\ shuffle), P25 (\evqa\ $k{=}20$), P26 (four-panel $k$-sweep generalization), P27 (text-only NQ ablation on \IV), P34-MULTIGOLD (multi-gold, with a pre-execution amendment restricting the population), and a REVIEW omnibus. Each note records its hypothesis and, where one applies, its decision rules verbatim. One registered element was not executed: the P34 note pre-registers the five-way classifier of App.~\ref{app:stats} for the multi-gold contrasts, and the runner defines but never calls it, so the multi-gold verdicts in \S\ref{sec:mechanism} rest on the paired-bootstrap intervals and McNemar tests we report rather than on a classifier label.

\noindent\textbf{Experiments run without a pre-registration note.} For completeness we state which experiments have no note rather than implying blanket coverage: P17c (oracle-rerank demo), P28 (\texttt{transformers} version regression), P29 (\QwenThree\ tiers), and the post-response runs P30, P32 and P33 (matched-corpus text-only controls), P31 (3B \infoseek), P35 (32B \infoseek) and P36 (32B \evqa). These reuse the frozen protocols of the pre-registered runs, and none of them defines a new decision rule. We flag them explicitly because the pre-registration guarantee applies only to the notes listed above.

\section{Subset Construction and Audits}
\label{app:subsets}
Both subsets are drawn with \texttt{Generator(PCG64(42))} and audited against the full released split at a $\pm 5$ pp gate. The \infoseek\ audit gates on image-filename prefix, positive-passage count and answer count, with question type and instruction prefix reported but not gated; the \evqa\ audit gates on image source, question type, positive-passage count and gold-answer length. Both PASS. The \evqa\ draw is taken directly from the released M2KR split; the \infoseek\ draw is taken from a fixed 2{,}000-question pool used in our earlier $k$-sweep.

\section{Statistical Methodology}
\label{app:stats}
Per-cell accuracy uses Wilson 95\% intervals ($z{=}1.96$). Between-cell deltas use a paired bootstrap over question ids with $B{=}10{,}000$ resamples, a PCG64 generator seeded at 42, and 2.5/97.5 percentiles. ``Separable'' throughout means the delta's 95\% interval excludes zero. Interaction quantities (a difference of two deltas, e.g.\ text-vs-image gaps or cross-model gap-of-gaps) resample question ids jointly across all four cells so the pairing is preserved.

\noindent\textbf{The five-way classifier, with its bands.} Given a delta, its interval, and a comparator anchor $A$, the verdict is: \textsc{flat} if the interval contains zero; otherwise \textsc{contradiction} if the delta's sign differs from $A$; otherwise, with $r = \Delta / A$, \textsc{amplified} if $r > 2.0$, \textsc{partial} if $r < 0.5$, and the same-magnitude band $0.5 \leq r \leq 2.0$ otherwise. We print the thresholds here so the classifier can be reconstructed from the paper alone.

\noindent\textbf{Label provenance.} The same-magnitude band was named \textsc{replication} in the originating pre-registration note (P17b), and the released result artifacts emit that string, apart from the 32B run, which emits \textsc{generalization}. Later notes and the version of this paper reviewed at ARR renamed it \textsc{generalization}, on the reasoning that most uses compare a \emph{different} reader or benchmark against the anchor rather than repeating the same cell. The thresholds were identical and no verdict differs under either name, but because a rename can read as a reinterpretation of a result, this version simply describes the band by its thresholds and reserves the word for the artifacts that emit it. Readers comparing the paper against released artifacts should expect to see \textsc{replication} there.

\noindent\textbf{Determinism.} Decoding is greedy, so the RNG seed does not affect generation; it fixes only the bootstrap resampling. Run-to-run reproducibility of the predictions themselves depends on GPU, batch composition and kernel backend. We disable the cuDNN SDP backend for stability, and we observed 96.6\% bit-identical predictions across a \texttt{transformers} 4.49-to-4.57 upgrade on the anchor cell, so a small amount of residual nondeterminism should be expected rather than exact bit-equality. Legacy independent-sample CIs are preserved as \texttt{*\_legacy\_indep} fields for reproducibility audits.

\section{Scoring Sensitivity}
\label{app:scoring}

\noindent\textbf{\evqa: does the matcher choice matter?} We recompute the $k{=}10$ first-vs-last delta under all three matchers. \Qwen: $+26.40$ $[+22.20,+30.60]$ under \texttt{orig} (substring), $+26.60$ $[+22.40,+30.80]$ under \texttt{wb} (word-boundary), $+17.80$ $[+13.80,+21.80]$ under \texttt{exact\_only}. \IV\ ($n{=}499$): $+23.25$ $[+19.04,+27.45]$, $+23.25$ $[+18.84,+27.66]$, $+19.24$ $[+15.23,+23.25]$. Every interval excludes zero and every classifier verdict is unchanged, so the effect is not an artifact of matcher leniency. Exact-equality scoring lowers the magnitude, as expected when free-form answers are penalized for surface form. Normalizer-mismatch quantification: 13 of 2{,}997 pred-side disagreements, 0 verdict flips.

\noindent\textbf{\infoseek: does strict scoring create the effect?} The concern runs the other way --- that the official exact-match rule marks verbose-but-correct answers wrong, and that those cluster at \emph{first}. We test it by re-scoring under lenient containment, in which a gold answer counts if it appears inside the prediction. Because the choice of lenient rule is arbitrary, we run three variants rather than privileging one: the official \infoseek\ normalizer with and without token-boundary anchoring, and the \evqa\ normalizer with anchoring. The gap \emph{widens} under every variant --- \Qwen\ from $+16.53$ to $+22.98$--$23.39$ pp, \QwenLarge\ from $+10.48$ to $+15.32$--$16.53$ pp --- so we report the spread rather than a point estimate, since the conclusion does not depend on which matcher is chosen. Taking the boundary-anchored official normalizer as the representative case, the intervals are $+22.98$ $[+19.15, +27.02]$ and $+15.32$ $[+11.69, +18.95]$. Strict scoring therefore understates the position effect.

\noindent\textbf{Naming.} This lenient rule is \emph{lexical containment}, optionally anchored to token boundaries. It is not BEM~\citep{bulian2022bem}, which is a learned BERT-based answer-equivalence model; no such model is loaded anywhere in this work, and we make no claim about how the effect would look under BEM itself. Both analyses in this appendix are produced by a single committed script over the released prediction files.

\section{Fallback-Gold Sensitivity}
\label{app:fallback_sens}
Paired-bootstrap clean-subset $\Delta$s for all six deltas after restricting to questions with no fallback-gold rule applied (\infoseek\ $n{=}354$; \evqa\ $n \in \{432, 433\}$). Maximum shift in any delta is 1.83 pp; classifier verdict labels are unchanged. Per-cell accuracies move more than the deltas do, since a shift shared by two cells cancels in their difference: across the thirteen \Qwen\ $\times$ \infoseek\ cells of the $k$-sweep (one at $k{=}1$, three at each of $k \in \{3,5,10,20\}$) every cell rises, by at most $5.2$ pp (largest: $k{=}3$ \emph{middle}, $40.32 \to 45.48$); the three $k{=}10$ cells rise by $2.9$--$4.1$ pp.

\section{Bucket Re-analysis Under Stricter Normalizer}
\label{app:bucket_renormalize}
\begin{figure}[h]
\centering
\includegraphics[width=\columnwidth]{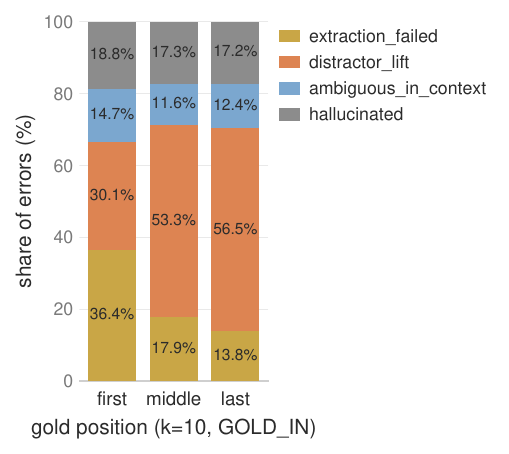}
\caption{\textbf{Failure-bucket distribution shifts with gold position.} Per-cell breakdown of P17 \infoseek\ \Qwen\ error rows at $k{=}10$ ($n{=}962$; 272 first, 336 middle, 354 last). As gold moves first $\to$ last, the modal failure mode flips from \texttt{extraction\_failed} ($36.4\% \to 13.8\%$) to \texttt{distractor\_lift} ($30.1\% \to 56.5\%$).}
\label{fig:buckets}
\end{figure}

We re-bucket all 962 P17 \infoseek\ \Qwen\ error rows at $k{=}10$ (272 first + 336 middle + 354 last) under a normalizer matching the answer-scoring normalizer. Result: at $k{=}10$ no error changes bucket and no bucket share shifts. Across the full sixteen-cell re-bucketing (5{,}004 rows, P17 and P17b), 3 errors flip and the maximum per-cell shift in any bucket is 0.3 pp. The qualitative pattern in Section~\ref{sec:mechanism} survives normalizer choice. Figure~\ref{fig:buckets} shows the bucket distribution by gold position.

\section{Multi-Gold Construction and Caveats}
\label{app:multigold_detail}

\noindent\textbf{Population.} Of the 496 \infoseek\ anchor questions, 142 have no positive in the \preflmr\ top-50 (fallback gold, excluded), 114 have exactly one, and 240 have two or more together with at least eight non-positive distractors. A pre-execution amendment further restricted the analysis population to questions where the answer string is independently present in \emph{both} chosen golds under the official normalizer, since a second gold that does not contain the answer confirms the hypothesis trivially. That check returned 240 of 240, so the amendment did not change the population. The note also pre-registered a sensitivity slice restricted to the 228 String/Time questions, for which substring answer-presence is most reliable; every verdict is unchanged on that slice (C4$-$C1 $+3.07$ $[-0.44, +6.58]$; C5$-$C3 $+7.46$; C1$-$C5 $+8.77$). A second gold at slot 4 rather than slot 9 is worth no more once slot 0 is gold: C6$-$C1 is $+0.42$ pp $[-2.92, +4.17]$. The two golds are the first two positives by \preflmr\ rank; distractors are the top-50 minus the full positive set, exactly as in the anchor.

\noindent\textbf{Redundancy, quantified.} All 240 gold pairs share a Wikipedia title. Content token-Jaccard between the two golds has median $0.163$, p10--p90 $0.109$--$0.671$, and full range $0.076$--$1.000$. Two of the 240 pairs are byte-identical passages that the knowledge base lists under two ids, for which ``different chunks'' is false; we retain them for population integrity and note that removing them cannot change a null result. An earlier pre-registration note described the spread as a $0.12$--$0.68$ range; that was a p10--p90 interval and is restated as such here.

\noindent\textbf{Annotation noise.} The M2KR \infoseek\ split provides pseudo-gold passages selected automatically rather than human-verified oracles. Multi-gold results should be read with that in mind: the positives are noisy, and a benchmark with curated multi-hop evidence would test a different and stronger claim.

\noindent\textbf{Cache reuse.} The three single-gold cells are prompt-identical to the corresponding anchor cells restricted to these 240 questions, so their predictions are reused rather than recomputed (720 of 720 cache keys matched, 0 prediction mismatches). Only the three two-gold cells required fresh inference.

\section{Per-Position Cost of Removing the Image}
\label{app:vision_protection}

Detail for the corresponding paragraph of \S\ref{sec:text_ablation}. Paired-bootstrap differences (text-only minus multimodal) per position cell, over the shared question set, are given in Table~\ref{tab:vision_protection}.

\begin{table*}[t]
\centering
\footnotesize
\caption{\textbf{Per-position cost of removing the image.} Paired-bootstrap differences (text-only minus multimodal) for each position cell, over the shared question set. On \infoseek\ the \emph{middle} and \emph{last} cells absorb most of the loss on both reader families; on \evqa\ no cell moves separably.}
\label{tab:vision_protection}
\begin{tabular}{llll}
\toprule
Cell & first & middle & last \\
\midrule
\Qwen\ $\times$ \infoseek     & $-4.03$ {\scriptsize $[-7.26,-0.81]$} & $-15.12$ {\scriptsize $[-18.75,-11.49]$} & $-10.08$ {\scriptsize $[-13.71,-6.45]$} \\
\Qwen\ $\times$ \evqa         & $+2.80$ {\scriptsize $[-0.20,+5.80]$} & $-2.00$ {\scriptsize $[-5.40,+1.40]$}   & $-0.60$ {\scriptsize $[-4.00,+3.00]$}  \\
\IV\ $\times$ \infoseek       & $-3.02$ {\scriptsize $[-6.85,+0.81]$} & $-6.65$ {\scriptsize $[-10.08,-3.42]$}  & $-4.23$ {\scriptsize $[-7.46,-1.01]$}  \\
\bottomrule
\end{tabular}
\end{table*}

\noindent On \infoseek\ the middle and last cells absorb most of the loss, on both reader families. On \evqa\ no cell moves separably, so the protection reading is \infoseek-specific and we do not generalize it.

\noindent\textbf{Row-level overlap with image-at-end.} Let $S_{\text{end}}$ be the questions answered correctly under the image-at-start default but wrongly under image-at-end, and $S_{\text{none}}$ the same for image-removed (\Qwen\ $\times$ \infoseek). At \emph{middle}, $|S_{\text{end}}|{=}67$, $|S_{\text{none}}|{=}87$, and $66$ of the $67$ lie inside $S_{\text{none}}$ (Jaccard $0.750$ against a chance level of $0.31$). At \emph{last}, $|S_{\text{end}}|{=}53$, $|S_{\text{none}}|{=}72$, overlap $52$ (Jaccard $0.712$, chance $0.27$). Moving the image and deleting it therefore damage the same specific questions rather than merely similar numbers of them, which is what motivates reading them as one mechanism at two doses. This is a descriptive row-level observation on one cell; we did not design an experiment to test it.

\section{Image-at-End Post-Hoc Finding}
\label{app:image_end_posthoc}
\begin{figure}[h]
\centering
\includegraphics[width=\columnwidth]{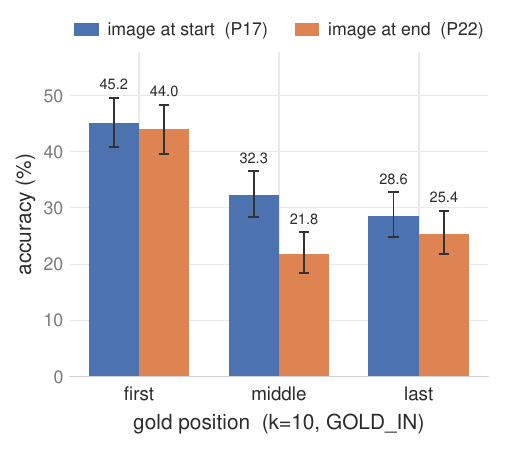}
\caption{\textbf{Image-token proximity is not the primary driver of first-vs-last.} Per-cell accuracy under image-at-start (P17 default) vs.\ image-at-end. First-vs-last gap preserved ($1.12\times$ anchor); middle/last ordering inverts under image-at-end (App.~\ref{app:image_end_posthoc} discusses).}
\label{fig:image_at_end}
\end{figure}

Under image-at-end, middle accuracy drops disproportionately (21.77\% vs.\ 32.26\% under image-at-start, $-10.48$ pp), while last drops less (25.40\% vs.\ 28.63\%, $-3.23$ pp). The middle-vs-last ordering inverts: $\Delta(\text{middle}-\text{last})$ flips from $+3.63$ pp at image-start to $-3.63$ pp at image-end (paired-bootstrap CI $[-6.05, -1.41]$ excludes zero). Image position therefore exerts a mild recency-style protection on adjacent passages but is too small to override the first-position prior, consistent with proximity playing a secondary modulating role on the middle/last contrast while the first-vs-last anchor remains driven by text-position. Figure~\ref{fig:image_at_end} compares the two templates.

\section{Distractor-Shuffle Cross-Tabulation (\infoseek)}
\label{app:p23_shuffle_fig}
\begin{figure}[h]
\centering
\includegraphics[width=\columnwidth]{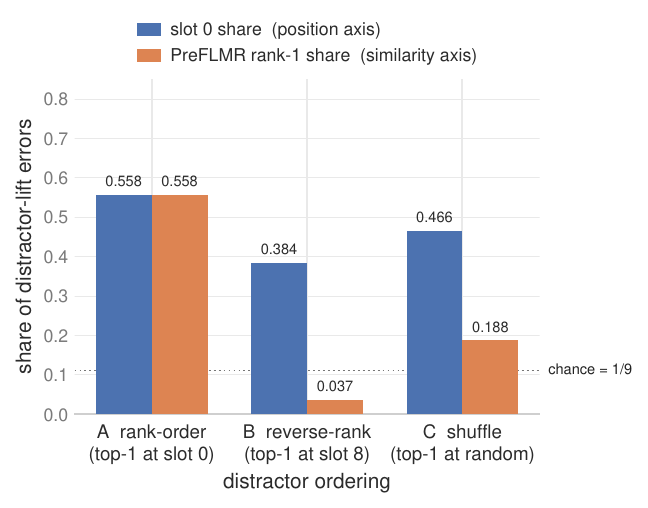}
\caption{\textbf{Reader follows prompt slot, not retrieval rank} (\infoseek). Distractor-lift error breakdown for the three shuffle cells (gold fixed at slot 9). Slot-0 selection (blue) dominates in dissociable cells B+C; rank-1 selection (orange) runs at the dashed baseline. Numbers are in Table~\ref{tab:shuffle}; the \evqa\ replication is in App.~\ref{app:p24_shuffle_evqa}.}
\label{fig:distractor_shuffle}
\end{figure}

Figure~\ref{fig:distractor_shuffle} plots the \infoseek\ cross-tabulation reported numerically in Table~\ref{tab:shuffle}: for each shuffle cell, the share of distractor-lift errors that selected the passage at slot 0 against the share that selected the passage at \preflmr\ rank 1. Disclosure: the pre-registration's verdict picker averages over all three cells, including the degenerate Cell A in which slot 0 and rank 1 coincide, and on that average it returns \textsc{mixed-leaning-position}; restricted to the dissociable cells B and C it returns \textsc{position}. We report the B+C reading throughout, since Cell A cannot separate the two axes by construction, and the released artifact records both.

\section{Shuffle Replication on \evqa\ (P24)}
\label{app:p24_shuffle_evqa}
\begin{figure}[h]
\centering
\includegraphics[width=\columnwidth]{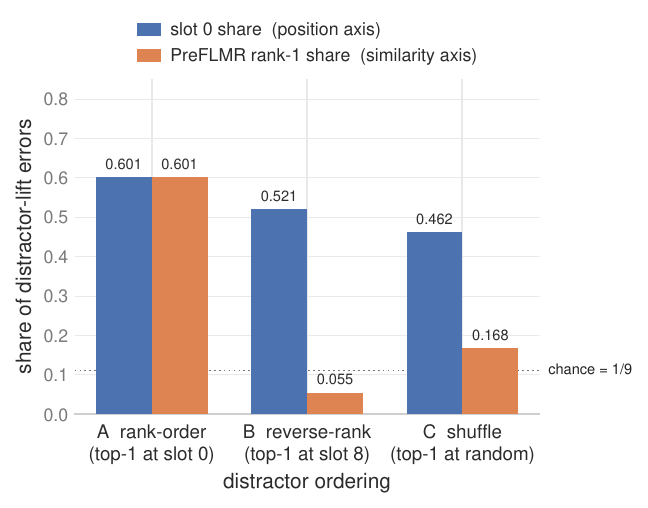}
\caption{\textbf{Shuffle ablation replicates on \evqa.} Distractor-lift error breakdown for the three shuffle cells on \Qwen\ $\times$ \evqa\ gold-at-last ($n{=}500$). Same protocol as Figure~\ref{fig:distractor_shuffle}; slot-0 dominance is stronger ($4.42\times$ vs $3.83\times$ chance baseline of $1/9$).}
\label{fig:p24_cross_tab}
\end{figure}

Figure~\ref{fig:p24_cross_tab} shows the cross-tabulation for the P24 shuffle replication on \Qwen\ $\times$ \evqa\ gold-at-last ($n{=}500$). The qualitative pattern is identical to the \infoseek\ result (Figure~\ref{fig:distractor_shuffle}): slot-0 anchoring dominates in dissociable cells B+C ($4.42\times$ chance, vs $3.83\times$ on \infoseek), while rank-1 selection sits at chance ($1.00\times$, vs $1.01\times$ on \infoseek). Aggregate-accuracy comparisons across the three cells against the deployment default are audited in App.~\ref{app:shuffle_operational_audit}.

\section{Full $k$-sweep on Six Reader\texorpdfstring{$\times$}{ x }Benchmark Cells (P26 + P29 Tier 2)}
\label{app:p26_ksweep}
Table~\ref{tab:p26_full} reports per-cell accuracies for the 6-panel $k$-sweep summarized in Figure~\ref{fig:p29_6panel}. The pattern is consistent across all six cells: first decays gracefully, middle and last drop steeply and collapse together; no recency rebound at $k{=}20$ on five of six cells. The sixth cell (\QwenThree\ $\times$ \evqa) shows a separable last $>$ middle effect at $k{=}10$ and $k{=}20$, discussed in \S\ref{sec:main} and App.~\ref{app:p29_qwen3vl}. Small middle-vs-last inversions occur on twelve sub-cells across the sweep, every one is within CI noise except the \QwenThree\ $\times$ \evqa\ cells discussed in \S\ref{sec:main}, the largest of the remainder being \IV\ \infoseek\ $k{=}3$ at $+2.82$ pp. \IV\ $\times$ \evqa\ incurs OOM exclusions on large prompts at every $k$, giving per-cell $n$ between 486 and 499 (most affected at $k{=}20$: $n \in \{486, 487, 487\}$ for first/middle/last after 13--14 exclusions per position cell); all other cells use the full subsets, $n{=}496$ for \infoseek\ and $n{=}500$ for \evqa.

\begin{table}[h]
\centering
\footnotesize
\caption{\textbf{Full $k$-sweep accuracies (first/middle/last \%; one prompt at $k{=}1$) across six reader$\times$benchmark cells.}}
\label{tab:p26_full}
\resizebox{\columnwidth}{!}{%
\begin{tabular}{lccccc}
\toprule
Reader$\times$Benchmark & $k{=}1$ & $k{=}3$ & $k{=}5$ & $k{=}10$ & $k{=}20$ \\
\midrule
\Qwen $\times$ \infoseek      & 54.4 & 48.0/40.3/41.1 & 47.0/37.3/34.7 & 45.2/32.3/28.6 & 43.5/26.6/26.0 \\
\Qwen $\times$ \evqa          & 70.2 & 65.0/53.8/50.0 & 64.0/48.6/42.2 & 63.2/38.6/36.8 & 62.8/33.0/34.0 \\
\IV   $\times$ \infoseek      & 50.2 & 44.4/28.8/31.7 & 43.8/27.4/28.0 & 40.1/23.8/22.2 & 38.7/20.6/19.8 \\
\IV   $\times$ \evqa          & 69.7 & 63.2/45.9/45.1 & 59.5/42.3/40.0 & 59.1/36.5/35.9 & 54.9/29.4/29.6 \\
\QwenThree $\times$ \infoseek & 58.7 & 55.0/41.5/40.7 & 52.2/37.1/38.9 & 51.6/33.9/35.5 & 50.4/31.9/32.9 \\
\QwenThree $\times$ \evqa     & 72.2 & 67.4/52.4/53.4 & 66.6/49.4/50.6 & 65.6/44.4/47.4 & 66.4/40.8/46.0 \\
\bottomrule
\end{tabular}}
\end{table}

\section{Text-only NQ-Open Ablation on \IV\ (P27)}
\label{app:p27_text_internvl3}
We run the NQ-Open text-only protocol on \IV\ using the identical 500-question subset from P21 (sampled from \citet{liu2024lost}'s 2{,}655 paragraph-answer queries, PCG64 seed 42). Per-cell accuracies (Wilson 95\% CI): first $55.00\%$ $[50.62, 59.31]$; middle $51.20\%$ $[46.83, 55.56]$; last $49.80\%$ $[45.44, 54.17]$. Paired-bootstrap deltas ($B{=}10{,}000$, seed 42): first$-$last $+5.20$ pp $[+1.20, +9.40]$; first$-$middle $+3.80$ pp $[+0.00, +7.60]$; middle$-$last $+1.40$ pp $[-2.00, +5.00]$ (CI includes zero: middle and last are flat, as on \Qwen, whose P21 run on the same subset gives first $63.60\%$, middle $55.40\%$, last $56.20\%$, i.e.\ first$-$last $+7.40$ pp $[+3.80, +11.00]$). This establishes that primacy is present for \IV\ on a second corpus and a different task format.

\section{\QwenThree\ Multimodal Replication (P29 Tiers 1, 2)}
\label{app:p29_qwen3vl}
We replicate the full position-effect protocol on \QwenThree-Instruct ($n{=}496$ \infoseek, $n{=}500$ \evqa, subsets identical to P17/P20). Tier 1 establishes the $k{=}10$ row of Table~\ref{tab:replication}: first-vs-last $+16.1$ pp $[+12.5, +20.0]$ on \infoseek\ ($0.98\times$ anchor; first 51.61\%, middle 33.87\%, last 35.48\%) and $+18.2$ pp $[+14.2, +22.2]$ on \evqa\ ($1.10\times$; first 65.60\%, middle 44.40\%, last 47.40\%). Both fall in the same-magnitude band of the pre-registered classifier. Tier 2 extends to $k \in \{1, 3, 5, 20\}$; per-cell accuracies are plotted in the two right-column panels of Figure~\ref{fig:p29_6panel}. Pre-registered verdicts for the recency-rebound test on $k{=}20$:
\begin{itemize}[noitemsep,topsep=2pt]
\item \QwenThree\ $\times$ \infoseek\ at $k{=}20$: $\Delta(\text{last}-\text{middle}) = +1.0$ pp, paired-bootstrap CI $[-1.0, +3.0]$. No recency rebound.
\item \QwenThree\ $\times$ \evqa\ at $k{=}20$: $\Delta(\text{last}-\text{middle}) = +5.2$ pp, paired-bootstrap CI $[+2.6, +8.0]$. \emph{Separable last $>$ middle} (CI excludes zero positively; see \S\ref{sec:main} for framing). At $k{=}10$ the same effect is present at $+3.0$ pp $[+0.2, +5.8]$.
\end{itemize}
This is the only cell of six showing a separable last $>$ middle effect in the default protocol.

\section{Text-only Ablation on \QwenThree\ (P29 Tier 3)}
\label{app:p29_text_qwen3vl}
A third NQ-Open text-only run on \QwenThree-Instruct (same 500-question subset as P21 and P27). Per-cell accuracies: first $62.0\%$, middle $59.2\%$, last $55.6\%$. Paired-bootstrap deltas: first$-$last $+6.4$ pp $[+3.2, +9.6]$; middle$-$last $+3.6$ pp $[+1.2, +6.2]$ (CI excludes zero positively; \QwenThree\ shows separable middle $>$ last in text mode, unlike \Qwen\ and \IV, which are flat). The text-mode middle $>$ last on \QwenThree\ inverts in multimodal \evqa\ to last $>$ middle by $5.2$ pp, the sign reversal reported in \S\ref{sec:main}; note that this contrast concerns the \emph{shape} of the middle/last tail rather than the magnitude of the first-vs-last gap, and it is the only cell of six where that tail separably inverts.

\section{Operational Audit of Shuffle Interventions}
\label{app:shuffle_operational_audit}

The body uses the Cell A / Cell B / Cell C cross-tabulation to establish that the reader follows prompt slot, not retrieval rank (the mechanism question). Here we report the corresponding aggregate-accuracy comparisons, which speak to a different question: does any shuffle variant improve on the deployment default (Cell A = rank-order, the configuration a normal pipeline would produce)?

\noindent\textbf{\infoseek\ (\Qwen, gold-at-last, $n{=}496$).} Per-cell accuracies (Wilson 95\% CI): A $28.83\%\ [25.02, 32.97]$; B $31.25\%\ [27.33, 35.46]$; C $28.02\%\ [24.25, 32.13]$. Paired-bootstrap deltas ($B{=}10{,}000$, seed=42): $\Delta(B-A) = +2.42$ pp $[-0.60, +5.65]$ (not separable); $\Delta(B-C) = +3.23$ pp $[+0.40, +6.05]$ (separable); $\Delta(C-A) = -0.81$ pp $[-3.43, +1.81]$ (flat).

\noindent\textbf{\evqa\ (\Qwen, gold-at-last, $n{=}500$).} Per-cell accuracies: A $36.80\%$ (matches the $k{=}10$ \evqa\ last cell exactly, confirming protocol determinism); B $33.60\%$; C $34.20\%$. Paired-bootstrap deltas: $\Delta(B-A) = -3.20$ pp $[-6.60, +0.00]$ (flat, point-negative); $\Delta(B-C) = -0.60$ pp $[-3.80, +2.40]$ (flat); $\Delta(C-A) = -2.60$ pp $[-5.80, +0.60]$ (flat).

\noindent\textbf{Reading.} The mechanism evidence (Cell B+C slot-0 dominance at $3.83$--$4.42\times$ chance, rank-1 at chance) is robust and benchmark-independent. The aggregate-accuracy story is different: on neither benchmark does any shuffle variant separably improve on the deployment default. The $\Delta(B-C) = +3.23$ pp on \infoseek\ that might initially read as ``demoting rank-1 helps'' is driven partly by Cell C dipping $\sim 0.8$ pp below default by chance, and disappears against the operationally relevant Cell A baseline.

\noindent\textbf{Scope of the test.} Cell B is an extreme intervention: it reverses all rankings, not just demotes rank-1. A surgical variant (``move rank-1 to slot 1 only, leave the rest alone'') was not tested and could plausibly behave differently. We do not recommend any training-free position-based intervention in the body of the paper.

\section{Oracle Reranking Visualization}
\label{app:oracle_rerank}
\begin{figure}[h]
\centering
\includegraphics[width=\columnwidth]{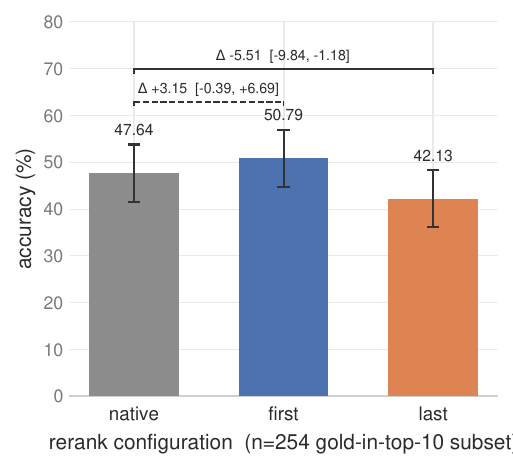}
\caption{\textbf{Oracle reranking does not separably help on a frozen reader.} \Qwen\ on $n{=}254$ gold-in-top-10 \infoseek\ subset. Solid bracket: paired-bootstrap CI excludes zero. Dashed bracket: CI includes zero. Forcing gold to last yields a separable loss; forcing gold to first yields a directional gain that does not clear the noise floor at this sample size.}
\label{fig:oracle_rerank}
\end{figure}

Figure~\ref{fig:oracle_rerank} shows the oracle-rerank result described in \S\ref{sec:mitigations}. Paired-bootstrap intervals on the $n{=}254$ subset: first$-$native $+3.15$ pp $[-0.39, +6.69]$; last$-$native $-5.51$ pp $[-9.84, -1.18]$. The first-vs-last contrast on the same subset is $+8.66$ pp $[+3.54, +14.17]$, separable and consistent with Table~\ref{tab:replication}.

\section{MMR Unit Tests}
\label{app:mmr}
8/8 PASS on synthetic ground-truth ordering at $\lambda \in \{0, 0.5, 1\}$. The unit tests fix the expected selection order at the extremes ($\lambda{=}1$ reduces to pure relevance, $\lambda{=}0$ to pure diversity) and at $\lambda{=}0.5$, and check that the first pick is the arg-max query similarity regardless of $\lambda$.

\begin{figure}[h]
\centering
\includegraphics[width=\columnwidth]{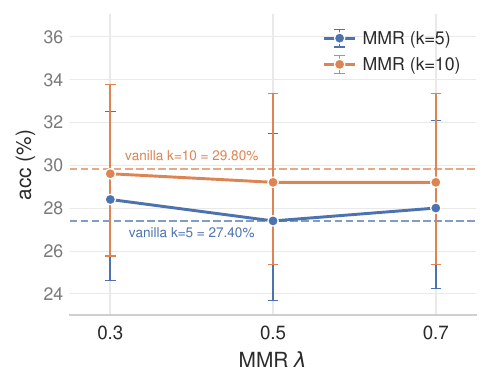}
\caption{\textbf{Diversification doesn't help.} MMR at three $\lambda$ values fails to improve over vanilla top-$k$ at $k{=}5$ or $k{=}10$. Every paired-bootstrap CI on $\Delta$(MMR$-$vanilla) overlaps zero.}
\label{fig:p19}
\end{figure}

\section{Distractor-Pool Determinism}
\label{app:distractor}
We confirm bit-identical $d_0, \ldots, d_{k-2}$ across \{first, middle, last\} cells per qid. We also verify P24 Cell A predictions are bit-identical to the P20 gold-at-last cell (cache reuse 500/500, distractor SHA256 hash failures = 0).

\section{Qualitative Error Samples}
\label{app:qualitative}
A sixty-row sample (five rows per bucket $\times$ four buckets $\times$ three cells) of P17 \infoseek\ \Qwen\ errors ships with the code release at \codeurl, under \texttt{results/\allowbreak p17/\allowbreak qualitative\_examples.md}.

\section{Prompts and Decoding}
\label{app:prompts}

All three readers share one system prompt and one user-message template, defined once and imported by each runner. The system prompt, verbatim and identical across readers:

\begin{quote}\ttfamily\footnotesize
You are a visual question answering assistant. Use the retrieved context to answer the question. Respond with a short phrase only, no explanation.
\end{quote}

\noindent The user message, verbatim:

\begin{quote}\ttfamily\footnotesize
\{question\}\\[2pt]
Context:\\
{}[1] \{passage 1 content\}\\
-{}-{}-\\
{}[2] \{passage 2 content\}\\
\ldots\\
{}[k] \{passage k content\}
\end{quote}

\noindent Passages are 1-indexed and joined by a literal \texttt{-{}-{}-} rule on its own line. Passage text is the M2KR \texttt{passage\_content} field used verbatim, with no re-wrapping. Note that the two benchmarks differ here: \infoseek\ \texttt{passage\_content} already begins \texttt{title: \ldots\ content: \ldots}, whereas \evqa\ passages carry no title field and contribute raw content only. There is no explicit answer cue; the assistant turn is opened by the chat template.

\noindent\textbf{Decoding, identical for every cell in the paper.} Greedy (\texttt{do\_sample=False}), \texttt{max\_new\_tokens=32}, no beam search, no top-$p$, no repetition penalty, \texttt{bfloat16} weights. Because decoding is greedy, no sampling seed enters the predictions.

\noindent\textbf{Reader-specific image handling} --- the only place the three readers differ. \Qwen\ and \QwenThree\ receive the system prompt through the chat template and place the image-token block first in the user message. \IV\ has no per-call system slot in its \texttt{model.chat()} path, so the same system text is prepended to the user text and the image is marked with a leading \texttt{<image>\textbackslash n}. Preprocessing follows each model's own recommendation, so that neither is handicapped: \Qwen\ and \QwenThree\ resize to a 1024 vision-token budget (\texttt{min\_pixels} $=56{\times}56$, \texttt{max\_pixels} $=1024{\times}28{\times}28=802{,}816$ for \Qwen\ and $1024{\times}32{\times}32=1{,}048{,}576$ for \QwenThree, whose patches are 16 pixels), while \IV\ uses dynamic tiling of up to 12 tiles of $448{\times}448$ plus a thumbnail. Only \Qwen\ exposes the image-position toggle used for the image-at-end ablation (\S\ref{sec:mechanism}).

\noindent\textbf{Text-only mode} (\S\ref{sec:text_ablation}). The image content block is removed from the user message entirely and no image argument is passed to the processor; everything else, including the system prompt, the passage block and the decoding configuration, is unchanged. Empirically the resulting prompts are a median 611 tokens shorter on \infoseek\ and 272 on \evqa\ than their multimodal counterparts on the same $(\text{question}, \text{passage ordering})$ pairs, bounded above by each reader's vision budget.

\noindent Because prompt construction and decoding are identical across position cells by construction, the position effect cannot be a prompt-formatting or sampling artifact.

\section{Compute and Reproducibility}
\label{app:compute}
H100 80GB; \texttt{bfloat16} weights; \texttt{torch} 2.11. P17--P27 (\Qwen\ and \IV\ experiments) were run under \texttt{transformers} 4.49.0. P29 (\textsc{Qwen3-VL} experiments; see App.~\ref{app:p29_qwen3vl}) was run under \texttt{transformers} 4.57.0, which is the minimum version supporting Qwen3-VL. We verified that the library upgrade introduces a small accuracy drift on \Qwen\ (+0.81 pp on the P17 anchor cell, with 96.6\% bit-identical predictions); this is well below our smallest reported effect size and does not flip any pre-registered verdict. We accordingly retain \texttt{transformers} 4.49.0 as authoritative for the P17--P27 numbers reported throughout the paper and use \texttt{transformers} 4.57.0 only for Qwen3-VL. The post-response runs P30--P36 ran under \texttt{torch} 2.11.0+cu128; P35 and P36 record \texttt{transformers} 4.49.0 in their released logs, and P30--P34 record no library version, inheriting the repository pin of \texttt{transformers} 4.49.0. Both the cu130 and cu128 CUDA builds of \texttt{torch} 2.11 appear across the experiment set, which is one source of the residual nondeterminism quantified in App.~\ref{app:stats}. Total session compute is approximately \$3--\$5 USD plus \$0.40 for the P21 text-only ablation, \$0.50 for the P24 shuffle replication, \$0.50 for the P25 $k{=}20$ replication, \$5 for the P26 four-panel $k$-sweep, \$0.50 for the P27 InternVL3 text-only ablation, and \$1--\$7 for the P29 Qwen3-VL experiments (tier-dependent; see App.~\ref{app:p29_qwen3vl}).

\end{document}